\documentclass[letterpaper]{article} 
\pdfoutput=1
\usepackage{aaai24}  
\usepackage{times}  
\usepackage{helvet}  
\usepackage{courier}  
\usepackage[hyphens]{url}  
\usepackage{graphicx} 
\usepackage{natbib}  
\usepackage{caption} 
\usepackage{algorithm}
\usepackage{algorithmic}
\usepackage{booktabs}
\usepackage{newfloat}
\usepackage{listings}
\usepackage{amsmath}
\usepackage{amssymb}
\usepackage{pifont}
\usepackage{xfrac}
\usepackage{microtype}
\usepackage{hyperref}
\hypersetup{
    colorlinks=true,
    linkcolor=red,
    filecolor=magenta,      
    citecolor=black
}
\newcommand{\cmark}{\ding{51}}%
\newcommand{\xmark}{\ding{55}}%

\usepackage{enumitem}
\usepackage{multirow}
\usepackage{colortbl}
\usepackage{microtype}
\usepackage{comment}
\usepackage[hang,flushmargin]{footmisc}

\usepackage{amsfonts}

\setlist[itemize]{align=parleft,left=0pt,topsep=1mm,itemsep=0mm,parsep=1mm}

\definecolor{azure(colorwheel)}{rgb}{0.0, 0.5, 1.0}
\definecolor{R5}{rgb}{0.0, 0.7, 0.1}
\definecolor{yw}{rgb}{0.01176, 0.5490, 0.5490}
\definecolor{R123}{rgb}{0.36, 0.54, 0.66}
\definecolor{R1234}{rgb}{0.7, 0.75, 0.71}
\definecolor{applegreen}{rgb}{0.55, 0.71, 0.0}
\definecolor{R132}{rgb}{0.0, 0.0, 1.0}
\definecolor{postechred}{rgb}{0.784, 0.003, 0.313}
\definecolor{gu}{rgb}{0.5460, 0.1755, 0.2766}
\definecolor{ballblue}{rgb}{0.13, 0.67, 0.8}
\definecolor{cornellred}{rgb}{0.7, 0.11, 0.11}
\definecolor{darkcyan}{rgb}{0.0, 0.55, 0.55}
\definecolor{CuGray}{gray}{0.9}
\definecolor{airforceblue}{rgb}{0.36, 0.54, 0.66}
\definecolor{rev}{rgb}{0.784, 0.003, 0.313}
\definecolor{pink}{cmyk}{0, 0.7808, 0.4429, 0.1412}
\definecolor{amethyst}{rgb}{0.6, 0.4, 0.8}
\definecolor{black}{rgb}{0.0, 0.0, 0.0}
\definecolor{tb3_yellow}{rgb}{0.996, 1.0, 0.6}
\definecolor{R123}{rgb}{0.980, 0.8, 0.604}
\definecolor{R512}{rgb}{0.972, 0.6, 0.6}
\definecolor{dimgray}{rgb}{0.41, 0.41, 0.41}
\definecolor{R3}{rgb}{0.8, 0.25, 0.33}
\definecolor{bleudefrance}{rgb}{0.19, 0.55, 0.91}
\definecolor{R6}{rgb}{0.265, 0.445, 0.765}
\definecolor{blue(ryb)}{rgb}{0.01, 0.28, 1.0}
\definecolor{R4}{rgb}{1.0, 0.49, 0.0}
\definecolor{Gray}{gray}{0.88}
\definecolor{green(ncs)}{rgb}{0.0, 0.62, 0.42}
\definecolor{brightpink}{rgb}{1.0, 0.0, 0.5}
\definecolor{alizarin}{rgb}{0.82, 0.1, 0.26}

\definecolor{kellygreen}{rgb}{0.3, 0.73, 0.09}

\newcolumntype{g}{>{\columncolor{CuGray}}c}
\newcolumntype{z}{>{\columncolor{CuGray}}l}

\renewcommand{\paragraph}[1]{\vspace{1mm}\noindent\textbf{#1.}\,\,}
\newcommand{\red}[1]{\textcolor{alizarin}{#1}}

\newcommand{\greencap}[1]{\textcolor{kellygreen}{#1}}

\usepackage{xspace}

\makeatletter
\def\@fnsymbol#1{\ensuremath{\ifcase#1\or *\or \dagger\or \ddagger\or
   \mathsection\or \mathparagraph\or \|\or **\or \dagger\dagger
   \or \ddagger\ddagger \else\@ctrerr\fi}}
\makeatother

\def\onedot{.\@\xspace}
\def\eg{\emph{e.g}\onedot} 
\def\ie{\emph{i.e}\onedot}

\newcommand{\ours}{FPRF}

\newcommand{\Sref}[1]{Sec.~\ref{#1}}
\newcommand{\Eref}[1]{Eq.~(\ref{#1})}
\newcommand{\Fref}[1]{Fig.~\ref{#1}}
\newcommand{\Tref}[1]{Table~\ref{#1}}

\newcommand{\be}{\begin{eqnarray}}
\newcommand{\ee}{\end{eqnarray}}
\newcommand{\bee}{\begin{eqnarray*}}
\newcommand{\eee}{\end{eqnarray*}}

\newcommand{\matrixb}{\left[ \begin{array}}
\newcommand{\matrixe}{\end{array} \right]}

\DeclareCaptionStyle{ruled}{labelfont=normalfont,labelsep=colon,strut=off} 
\floatstyle{ruled}
\newfloat{listing}{tb}{lst}{}
\floatname{listing}{Listing}
\title{FPRF: Feed-Forward Photorealistic Style Transfer of\\ Large-Scale 3D Neural Radiance Fields}
\author {
    GeonU Kim\textsuperscript{\rm 1},\quad
    Kim Youwang\textsuperscript{\rm 2},\quad
    Tae-Hyun Oh\textsuperscript{\rm 1,\rm 2,\rm 3}
}
\affiliations {
    \textsuperscript{\rm 1}Grad. School of Artificial Intelligence and 
    \textsuperscript{\rm 2}Dept. of Electrical. Engineering, POSTECH\\
    \textsuperscript{\rm 3} Institute for Convergence Research and Education in Advanced Technology, Yonsei University\\
    \texttt{\{gukim, youwang.kim, taehyun\}@postech.ac.kr}
}

\begin{document}

\maketitle

\begin{abstract}
We present \ours, a feed-forward photorealistic style transfer method for large-scale 3D neural radiance fields.
%
\ours\, stylizes large-scale 3D scenes with 
arbitrary, multiple style reference images without additional optimization while preserving multi-view appearance consistency.
%
Prior arts required tedious per-style/-scene optimization and were limited to small-scale 
3D scenes. 
\ours~efficiently stylizes large-scale 3D scenes by introducing a style-decomposed 3D neural radiance field, which inherits AdaIN's feed-forward stylization machinery, supporting arbitrary style reference images. 
%
%
Furthermore, \ours~supports multi-reference stylization with the semantic correspondence matching and local AdaIN, which adds diverse user control for 3D scene styles.
\ours~also preserves multi-view consistency by applying semantic matching and style transfer processes directly onto queried features in 3D space.
In experiments, we demonstrate that \ours~achieves favorable photorealistic quality 3D scene stylization for large-scale scenes with diverse reference images.  \textit{Project page: \href{https://kim-geonu.github.io/FPRF/}{https://kim-geonu.github.io/FPRF/}}
%
\end{abstract}

\section{Introduction}
\label{sec:intro}
Large-scale 3D scene reconstruction is a longstanding problem in computer vision and graphics~\cite{fruh2004automated,snavely2006photo,pollefeys2008detailed, agarwal2009rome, zhu2018very,tancik2022blocknerf}, which aims to build realistic 3D virtual scenes from a set of images.
Recently, Neural Radiance Fields~\cite{mildenhall2020nerf, barron2021mip, barron2022mip, fridovich2022plenoxels, jun2022hdr, muller2022instant, fridovich2023kplanes} and its large-scale extensions~\cite{tancik2022blocknerf, turki2022mega,zhenxing2022switch} 
have shown remarkable progress in modeling coherent and photorealistic outdoor 3D scenes, suggesting promising future directions, \eg, VR/AR applications.
%
In this work, we take a step further and focus on a task, 
photorealistic style transfer (PST) for large-scale 3D scenes, \ie, 3D scene PST.

The 3D scene PST task aims to 
transfer the visual styles of style reference images 
onto a large-scale 3D scene represented by a \emph{neural radiance field}. 
Within this objective, the resulting stylized output is expected to be photorealistic and preserve the geometric structure of the original scene.
Large-scale 3D scene PST has various applications,
where it can enrich virtual 3D spaces of XR applications and realistically augment existing real-world autonomous driving datasets~\cite{geiger2012kitti,cordts2016cityscapes}.

Recent studies~\cite{chen2022upst, zhang2023transforming} 
have developed methods for 3D scene PST 
and shown plausible visual qualities.
However, all of them 
%
require additional 
time-consuming learning and optimization stages even after scene reconstruction~\cite{chen2022upst}
or tedious per-style optimization steps to apply even just a single style to the scene~\cite{zhang2023transforming}.
More importantly, due to the aforementioned drawbacks, the previous 3D scene PST methods do not scale to large 3D scenes.
This motivates us to develop an efficient PST method for large-scale 3D scenes that stylizes the whole 3D scene without exhaustive and time-consuming per-scene or per-style optimization.

\begin{table}[t]
    \setlength\tabcolsep{2.0pt}
        \renewcommand{\arraystretch}{1.2}
    \small
    \begin{tabular}{l cccc}
        \toprule
        & \small{UPST-NeRF} & \small{StyleRF} & \small{LipRF} &  \small{\textbf{FPRF}}\\
        & \scriptsize(arXiv2022) & \scriptsize(CVPR2023) & \scriptsize(CVPR2023) & \scriptsize\textbf{(Ours)} \\ 
        \cmidrule{1-5}
        Photorealistic & \greencap{\cmark} & \red{\xmark}& \greencap{\cmark}& \greencap{\cmark}\\
        Feed-forward & \greencap{\cmark} & \greencap{\cmark}& \red{\xmark}& \greencap{\cmark}\\
        Multi-reference & \red{\xmark} & \red{\xmark}& - & \greencap{\cmark}\\
        Single-stage training & \red{\xmark} & \red{\xmark}& \red{\xmark}& \greencap{\cmark}\\
        \bottomrule
    \end{tabular}
    \caption{{FPRF} provides a photorealistic, feed-forward style transfer method for large-scale neural radiance fields, while supporting multiple style reference images in the stylization stage. Our model, {FPRF}, differs from competing methods, where it needs only an efficient single stage training, and it works with arbitrary style images in a feed-forward manner.}
    \label{tab:property}
\end{table}

       

In this work, we propose FPRF, 
an efficient and 
feed-forward PST for large-scale 3D scenes.
%
%
%
%
To implement a feed-forward
PST method for a large-scale 3D scene,
we employ the adaptive instance normalization (AdaIN) method, which has shown efficient and promising style transfer results on 
various tasks~\cite{huang2017adain,huang2018multimodal, gunawan2023modernizing, wang2020neural, aberman2020unpaired, segu20203dsnet, huang2022stylizednerf, liu2023stylerf}.
Specifically, we propose 
a photorealistic \emph{stylizable radiance field} consisting of a scene \emph{content} field and a scene \emph{semantic} field.
%
Given a large set of photos of the target scene, we first train a grid-based scene content field to embed the scene geometry and content features that can be later decoded to a large-scale radiance field of arbitrary styles.
The scene semantic field is trained together to match
proper local styles to the local scene.
%
After obtaining the scene content field, our FPRF stylizes the whole 3D scene via AdaIN,
that manipulates the scene content field with the 
style reference images' feature statistics in a feed-forward manner without any nuisance optimization.

In addition to efficiency, stylizing a large-scale scene requires dealing with diverse objects and contents that are hard to cover with a single reference image. 
Also, it is challenging to identify a single reference that can effectively encompass the entire semantics of a large-scale scene. 
Thus, extending the existing single reference-based methods \eg, \citet{chen2022upst}, is not straightforward,
due to diverse contents in the large-scale scene.
To overcome this challenge, we introduce a style dictionary module and style attention for efficiently retrieving the style matches of each local part of the 3D scene from a given set of diverse style references.
The proposed style dictionary consists of pairs of a local semantic code and a local style code extracted from the style references.
To form a compact style dictionary, we cluster similar styles and semantics and use the centroids of the clusters as the dictionary's elements, notably reducing the computational complexity of the style retrieval.
Using a style dictionary, we find semantic correspondences between the local semantic codes and the style semantic field.

Our experiments
demonstrate that
FPRF obtains superior stylization qualities
for both large/small-scale scenes 
with multi-view consistent
and semantically matched 
style transfer.
Furthermore, our model demonstrates 
versatility compared to the previous methods by stylizing 3D scenes with multiple style references, which is not feasible with other prior methods.
Our main contributions are summarized as follows:
\begin{itemize}
    \item We propose a stylizable radiance field where we can perform photorealistic style transfer in a feed-forward manner with an efficient single-stage training.
    \item We propose the style dictionary and its style attention for style retrieval, which allows us to deal with multiple style references efficiently.
    \item 
    To the best of our knowledge, 
    our work is the first multi-reference based 3D PST without any per-style optimization, which is scalable for large-scale 3D scenes. 
\end{itemize}

\section{Related Work}
\label{sec:related}
Our task relates to large-scale neural scene reconstruction and photorealistic style transfer for large-scale 3D scenes via neural feature distillation.

\paragraph{Large-scale neural 3D scene reconstruction}
Realistic 3D reconstruction of large-scale scenes
has been considered an important task,
which could be a stepping stone to achieve a comprehensive 3D scene understanding and immersive virtual reality.
Recently, a few seminal studies~\cite{tancik2022blocknerf,turki2022mega,zhenxing2022switch} tackled the task by leveraging 
advances in neural radiance fields~\cite{mildenhall2020nerf}, showing remarkable reconstruction quality for large-scale 3D scenes. 
%

Prior arts mainly focused on decomposing the large-scale scene into smaller parts, \ie, divide-and-conquer.
Block-NeRF~\cite{tancik2022blocknerf} manually divided
city-scale scenes into block-level, 
and Switch-NeRF~\cite{zhenxing2022switch} learns scene decomposition using a learnable gating network.
%
%
%
Although they can reconstruct large scenes by decomposition, they take a long time to train
and inference by using a structure composed of a large neural network~\cite{mildenhall2020nerf, barron2021mip}.
Instead of adopting time-consuming MLP-based architecture, 
we leverage lightweight and easy-to-learn K-planes~\cite{fridovich2023kplanes} for representing large-scale 3D scenes and build our system, FPRF.
Specifically, using an efficient blockwise decomposed K-planes representation, we learn not only the scene geometry and radiance field, but also learn per-3D point high dimensional features for the large-scale scene, which we call the \emph{stylizable radiance field}. 
%
With the stylizable radiance field
and our 
multi-view consistent MLP color decoder, FPRF embeds compatibility for stylizing large-scale scenes with any style reference images while preserving the high-fidelity reconstruction quality for the original scene even without post-optimization.

\paragraph{Photorealistic style transfer for 3D scenes}
3D scene style transfer aims to stylize 3D scene according to the style of the reference image while preserving multi-view consistency. 
Recently,
\citet{chiang2022stylizing, huang2022stylizednerf, fan2022unified, nguyen2022snerf, chen2022upst, zhang2022arf, liu2023stylerf, zhang2023transforming} combine 3D neural radiance field with style transfer. 
%

Among them, UPST-NeRF~\cite{chen2022upst} and LipRF~\cite{zhang2023transforming} tackled PST on 3D neural radiance field. 
UPST-NeRF constructs a stylizable 3D scene with a hyper network, which is trained on stylized multi-view images of a single scene.
They can stylize a trained scene in a feed-forward manner with arbitrary style. However, the model requires additional time-consuming (${>}10$ hrs.) per-scene optimization after scene reconstruction.
%
%
LipRF stylizes the reconstructed 3D scene with multi-view stylized images by 2D PST methods.
They achieved 3D PST by leveraging Lipschitz network~\cite{virmaux2018lipschitz} to moderate artifacts and multi-view inconsistency caused by 2D PST methods. 
However, LipRF requires a time-consuming iterative optimization procedure for every unseen reference style.
%
One of the artistic style transfer methods, StyleRF~\cite{liu2023stylerf} leverages the 3D feature field to stylize 3D scenes with artistic style.
They achieve feed-forward style transfer with distilled 3D feature field, however, it requires per-scene multi-stage training which spends more time than reconstruction (${>}5$hrs.). 
Our model 
mitigates aforementioned inefficiency by constructing stylizable 3D radiance field 
with only a single training stage spending about 1 hr.
Also, our work distinctively focuses on photorealistic stylization with the capability of referring multi-style references, while StyleRF focuses only on artistic stylization with a single reference.
%
\Tref{tab:property} compares the distinctiveness of our work with the prior work.

%




%

\section{Method}
In this section, we introduce FPRF, a feed-forward Photorealistic Style Transfer (PST) method for large-scale 3D scenes.
%
%
We first introduce a single-stage training of the stylizable radiance field using AdaIN (\Sref{sec:pst_adain}). 
%
We further describe the scene semantic field for stylizing large-scale scenes with multiple reference images (\Sref{sec:pst_semantic_matching}).

\begin{figure}[t]
    \centering
    \includegraphics[width=\linewidth]{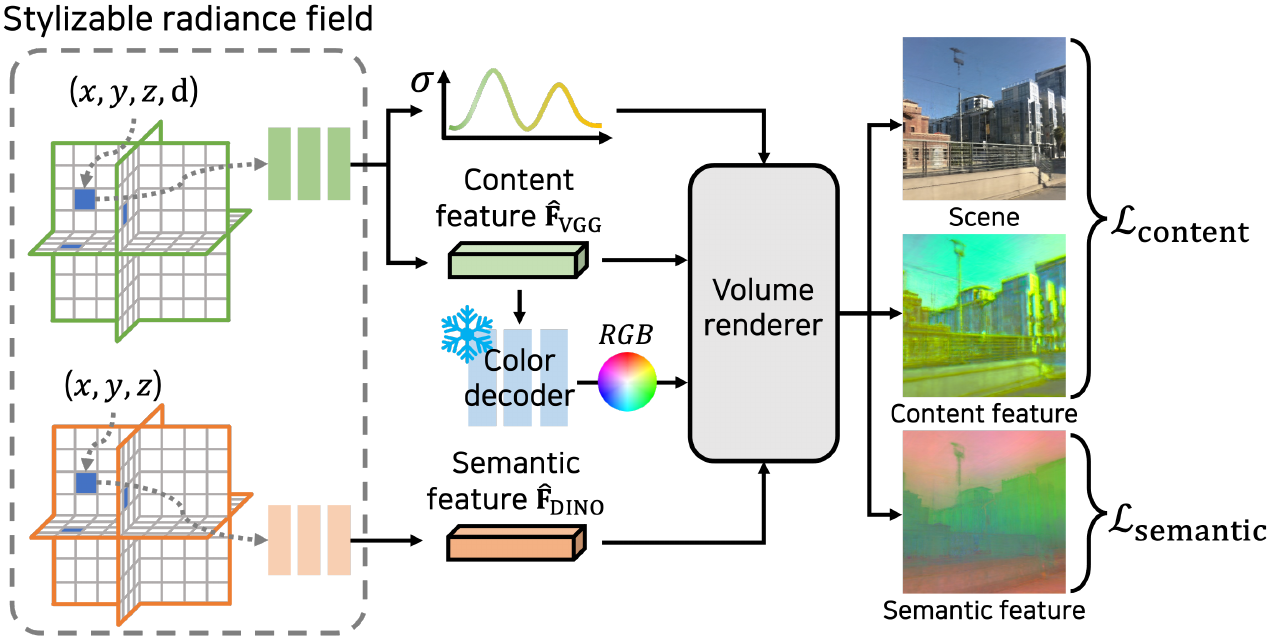}
    \caption{\textbf{FPRF training stage.} Given a set of scene images and corresponding VGG and DINO features, {FPRF} learns the stylizable radiance field. 
    Stylizable radiance field 
    embeds the geometry, radiance field, and semantic features of the scene. 
    Note that {FPRF} only needs the original scene images during training, not the stylized images, while it can take arbitrary style images in the stylization stage.}
    \label{fig:pipeline_train}
\end{figure}



\subsection{Large-scale Stylizable Radiance Field}
\label{sec:pst_adain}
Our goal is to construct a 3D large-scale radiance field that can be stylized with reference images in a feed-forward manner while achieving photorealistic results after stylization.
For the feed-forward style transfer, we employ adaptive instance normalization
(AdaIN)~\cite{huang2017adain}, which
%
is an efficient
style reference-based style transfer method,
where it linearly replaces
the content image's feature statistics with a given style image's feature as:
\begin{equation}
    \text{AdaIN}(\mathbf{F}(c), \mathbf{F}(s)) =\sigma_s\frac{(\mathbf{F}(c) - \mu_c)}{\sigma_c} + \mu_s,
\end{equation}
where
\(\mathbf{F}(c)\) and \(\mathbf{F}(s)\) are semantic features extracted from content and style images by a pre-trained feature encoder, \eg, a VGG encoder~\cite{simonyan2014very}. 
The feature statistics \(\mu_{*}\) and \(\sigma_{*}\) over spatial axes are the mean and standard deviation of the extracted features, respectively.
The AdaIN layer is favorable to enable feed-forward style transfer.
This property is particularly useful for dealing with 
large-scale 3D scenes and enables 
%
a fast and seamless stylization.
To leverage AdaIN's statistic-based style transfer to the 3D scene,
we propose to reconstruct a stylizable 3D neural radiance field by distilling 
the 2D features onto the field, called the stylizable radiance field.

\paragraph{Stylizable radiance field}
To build a multi-view consistent stylizable feature field, 
we apply multi-view bootstrapping~\cite{simon2017hand} to our domain.
We distill  
high-dimensional features obtained from 2D input images into neural feature fields that models the large-scale 3D scene. 
To represent a large-scale 3D scene, we extend 
K-planes~\cite{fridovich2023kplanes} with 
the block-composition manner~\cite{tancik2022blocknerf}, which is used for 
embedding volumetric scene geometry, radiance, and semantic features.
%
Specifically, we design the stylizable radiance field
with two tri-plane grids: scene \emph{content} field and scene \emph{semantic} field. 
The scene semantic field embeds
semantic features of a scene, which will be discussed later in \Sref{sec:pst_semantic_matching}.
The scene content field is responsible for embedding 
accurate scene geometry and appearance-related content features.
%
Given a 3D scene point position $\mathbf{x}{=}(x, y, z)$ and a ray direction vector $\mathbf{d}$ as inputs, our scene content field outputs
the density, and content feature of the query 3D point (see \Fref{fig:pipeline_train}).
%
%

The original NeRF 
computes a pixel's
RGB color $\hat{\mathbf{C}}(\textbf{r})$ by accumulating the color \(\hat{\mathbf{c}}_i\) and density \(\sigma_i\) of 
sampled 3D points
\(\mathbf{x}_i\) 
along the ray \textbf{r}.
Correspondingly,
we train to render
high-dimensional features
of 3D points in the scene
to a pixel value $\hat{\mathbf{F}}(\mathbf{r})$ by
accumulating
features \(\hat{\mathbf{f}}_i\)  
along the ray $\mathbf{r}$:
\begin{equation}
\hat{\mathbf{F}}(\mathbf{r}) = \sum^{K}_{i=1}\text{exp}\left( -\sum^{i-1}_{j=1}\sigma_j\delta_j\right)(1-\text{exp}(-\sigma_i\delta_i))\hat{\mathbf{f}}_i,
\label{eq:feature volume rendering}
\end{equation}
where \(\delta_i\) denotes the distance 
between the sampled point $\mathbf{x}_i$ and the
next sample point, \(\text{exp}(-\sum^{i-1}_{j=1}\sigma_j\delta_j)\) is a transmittance 
to the point \(\mathbf{x}_i\), 
and \(1-\text{exp}(-\sigma_i\delta_i)\) represents the absorption by \(\mathbf{x}_i\).
We distill 2D image features to a 3D scene by minimizing the error between the volume-rendered features and the 
2D features extracted from input images, which we call feature distillation.
To distill
highly-detailed features into the
3D scenes, we use the refined 2D features
by Guided Filtering~\cite{he2012guided} before distillation.

\begin{figure*}[t!]
    \centering
    \includegraphics[width=\linewidth]{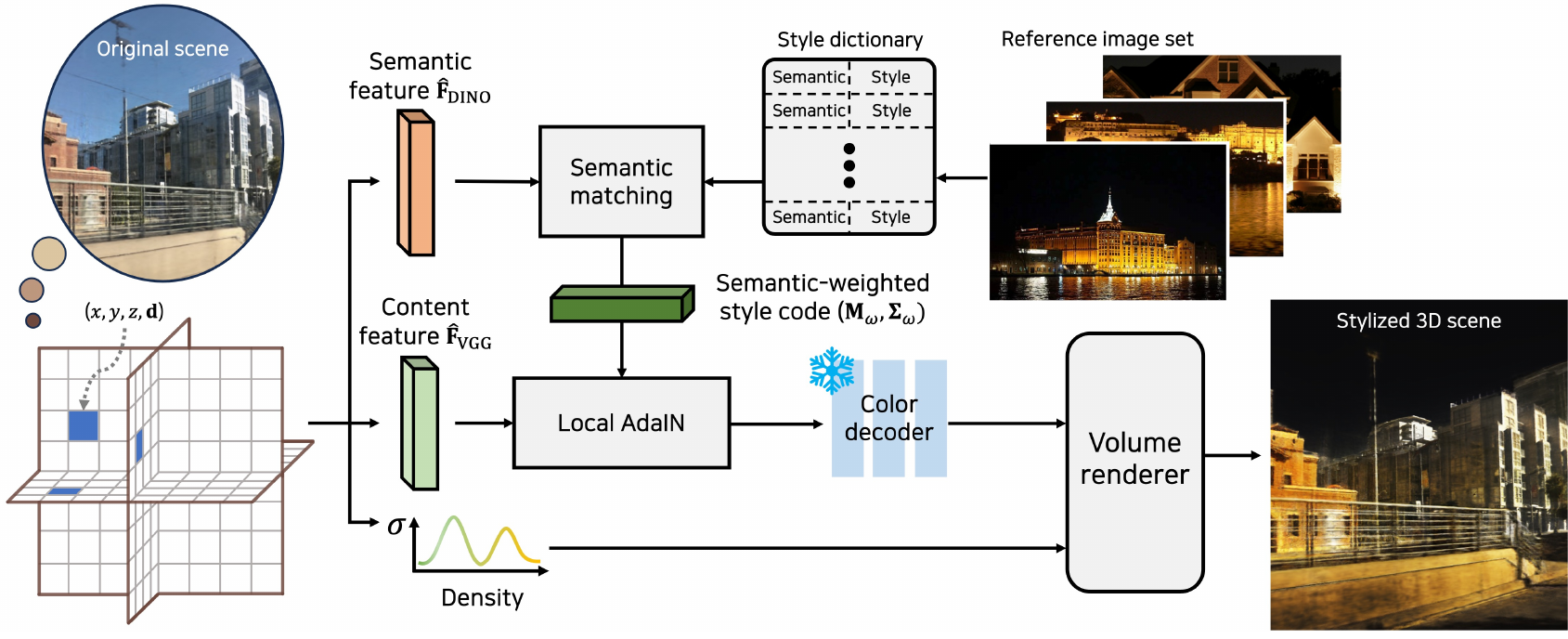} 
    \caption{\textbf{{FPRF} stylization stage}. Given the optimized 
    stylizable radiance field and the set of arbitrary reference images, we stylize the large-scale 3D scene via our novel semantic-aware local AdaIN. 
    We compose a style dictionary consisting of local semantic codes and local style codes pairs extracted from the clustered reference images. 
    Using semantic features from the stylizable radiance as a query, we find the corresponding local semantic features and retrieve the paired local style codes. 
    %
    %
    Using the retrieved semantic-style code pairs, we perform 
    semantic matching and local AdaIN,
    then finally render the stylized colors.
    }
    \label{fig:pipeline_inference}
\end{figure*}

\paragraph{Generalizable pre-trained MLP color decoder}
The output of the scene content field requires a separate decoder to decode the stylized feature into the color.
One may na\"ively train such a decoder scene-by-scene, which  suffers from a limited generalization to unseen colors 
and requires additional training for stylization.
In the following, we present a scene-agnostic and 
pre-trained MLP color decoder \(D_\text{VGG}\) compatible with AdaIN.
Specifically, 
\(D_\text{VGG}\)
transforms the distilled VGG features into 
colors.
To perform style transfer 
in a feed-forward manner, we pre-train \(D_\text{VGG}\) with 
the diverse set of content images~\cite{lin2014microsoft} and style images~\cite{nichol2016painter}, which enable \(D_\text{VGG}\) to be generalized to arbitrary style reference images.
For that, we employ content loss \(\mathcal{L}_c\) and the style loss \(\mathcal{L}_s\), similar to \citet{huang2017adain}:
$\mathcal{L}_{D_\text{VGG}} = \mathcal{L}_c + \lambda_{s}\mathcal{L}_s$.
The main differences with the training process of AdaIN are threefold.
First, we use the MLP architecture instead of the CNN upsampling layer inducing multi-view inconsistency.
Also, we employ features from the shallower layer, \texttt{ReLU2\textunderscore1} of VGGNet, that contain richer color information.
Furthermore,
the features from the input content images are upsampled to the pixel resolution
for per-pixel decoding.
%
When training the stylizable radiance field,
we fix the pre-trained $D_\text{VGG}$ 
so that it preserves the
knowledge about the distribution of VGG features from diverse images, which induces compatibility with AdaIN.


\paragraph{Training scene content field}
We train the scene content field by optimizing
the tri-plane grid features and MLP.
%
In detail, for the input 3D point $\mathbf{x}_i$ and the view direction vector $\mathbf{d}$, the grid features of the scene content field are decoded into the 3D point density $\sigma_i$ and the content feature (see the green grid and MLP in \Fref{fig:pipeline_train}).
We further decode the content feature into RGB values using the pre-trained $D_\text{VGG}$.
%
%
At the end, we render the content feature \(\hat{\mathbf{F}}_\text{VGG}(\textbf{r})\) and the scene color \(\hat{\mathbf{C}}(\textbf{r})\) via volume rendering (\Eref{eq:feature volume rendering}).

To perform AdaIN, we guide the content feature map ${\hat{\mathbf{F}}}_\text{VGG}$ to follow VGG feature distribution by feature distillation.
We distill the ground truth features \(\mathbf{F}_\text{VGG}(\mathbf{I})\) obtained from input images 
\(\mathbf{I}\) 
via pre-trained VGGNet, by minimizing the
error between \(\mathbf{F}_\text{VGG}(\mathbf{I})\) and the volume rendered features \({\hat{\mathbf{F}}}_\text{VGG}(\textbf{r})\).
Note that the ground truth VGG feature maps
are upsampled to pixel resolution with guided filtering.
Since the scene content field needs to reconstruct 
the accurate scene geometry and appearance, we compute the photometric loss for the volume rendered color \(\hat{\mathbf{C}}(\mathbf{r})\).
Also, typical regularization losses~\cite{fridovich2023kplanes} are employed.
The total loss function for training the scene content field is as below:
\begin{multline}
    \mathcal{L}_\text{content} = \sum_{\mathbf{r} \in \mathcal{R}} \lVert \hat{\mathbf{F}}_\text{VGG}(\mathbf{r}) - \mathbf{F}_\text{VGG}(\mathbf{I}, \mathbf{r})\rVert^2_2 \\ + 
    \lambda_{\text{RGB}}\sum_{\mathbf{r} \in \mathcal{R}} \lVert \hat{\mathbf{C}}(\mathbf{r}) - \mathbf{C}(\mathbf{I}, \mathbf{r})\rVert^2_2  + 
    \lambda_\text{reg}\mathcal{L}_\text{reg},
\label{eq:loss_color}
\end{multline}
where \(\mathcal{R}\) is the set of sampled rays in each training batch, and \(\mathbf{F}_\text{VGG}(\mathbf{I}, \mathbf{r})\) and \(\mathbf{C}(\mathbf{I}, \mathbf{r})\) denote ground truth VGG features and RGB values of the pixels correspond to the ray \(\mathbf{r}\in\mathcal{R}\). 

Note that we keep \(D_\text{VGG}\) frozen after its pre-train stage, \ie, \(D_\text{VGG}\) is fixed during the training stage of the scene content field.
This differs from StyleRF~\cite{liu2023stylerf}, which needs to fine-tune a CNN-based decoder for each scene.

\paragraph{Feed-forward stylization using the scene content field}
After training the scene content field,
we can perform PST with an arbitrary style image in the stylization stage.
In other words, we train the scene content field once and perform PST on a trained radiance field in a feed-forwards manner, \ie, without per-style or per-scene optimization.
%

Given a reference image 
\(\mathbf{I}_s\),
we start with extracting the VGG feature \(\mathbf{F}_\text{VGG}(\mathbf{I}_s)\).
Then we stylize 3D content features \(\hat{\mathbf{F}}_\text{VGG}\) as \( \sigma_s\sfrac{(\hat{\mathbf{F}}_\text{VGG} - \mu_c)}{\sigma_c} {+} \mu_s\), where \(\mu_s\) and \(\sigma_s\) are the mean and standard deviation of \(\mathbf{F}_\text{VGG}(\mathbf{I}_s)\).
By decoding the stylized content features to RGB values using the pre-trained color decoder \(D_\text{VGG}\), 
we can render a stylized 3D scene.
The mean and standard deviation of \(\hat{\mathbf{F}}_\text{VGG}(\mathbf{r})\) are keep tracked with the moving average during training~\cite{liu2023stylerf}. 

%


\subsection{Multi-reference image 3D Scene PST via Semantic matching and Local AdaIN}
\label{sec:pst_semantic_matching}
With the trained scene content field and AdaIN, we can efficiently transfer the style of images to the 3D radiance field. 
However, it often fails to produce satisfying results when it comes to large-scale 3D scenes:
AdaIN only allows a single reference image which often cannot cover all components in the large-scale scene.
%
To overcome this limitation, we propose a 
multi-reference based 3D scene PST 
by semantically matching the radiance field and multiple reference images.
As shown in the \Fref{fig:pipeline_inference}, the process involves two steps.
First, we compose a style dictionary containing local semantic-style code pairs
obtained from 
semantically 
clustered reference images. 
Then, we perform a semantic-aware style transfer 
by leveraging the semantic correspondence between the 3D scene and each element of the composed style dictionary.

\paragraph{Reference image clustering}
To stylize a 3D scene with multiple style reference images,
we consider a set of reference images, $\mathcal{I}_{s}{=}\{\mathbf{I}_{s}^{i}\}_{i=1,\dots,N}$, where $N$ denotes the number of reference images we use.
We compose a compact style dictionary $\mathcal{D}$ with the reference images by clustering them with similar styles and semantics.
We first extract semantic feature maps to cluster the reference images according to semantic similarity.
We employ DINO~\cite{caron2021emerging} as a semantic feature encoder, which can be generalized to various domains by being trained on large-scale datasets in a self-supervised manner.
We then apply K-means clustering to the extracted semantic feature map \(\mathbf{F}_\text{DINO}(\mathbf{I}_s^i)\), 
and obtain semantically correlated $M$ number of clusters \(\mathcal{S} {=}\{\mathbf{S}^{ij}\}_{i=1,\dots,N}^{j=1,\dots,M} \) from each reference image $\mathcal{I}_{s}$.
We obtain local style codes from the clusters by extracting another feature map \(\mathbf{F}_\text{VGG}(\mathbf{I}_s^i) \) from each reference image with VGGNet~\cite{simonyan2014very}.
Then we obtain the local style code, mean \(\mu_s^{ij}\) and standard deviation \(\sigma_s^{ij}\), from VGG features \(\mathbf{F}_\text{VGG}(\mathbf{I}_s^{ij}) \in \mathbf{S}^{ij}\) assigned to each cluster.
%
The centroid \(\bar{\mathbf{f}}_\text{DINO}(\mathbf{I}_s^{ij})\) of each clustered semantic feature and the assigned local style code (\(\mu_s^{ij}, \sigma_s^{ij}\))  compose a key-value pair for the style dictionary \(\mathcal{D}\), as \(\mathcal{D}_{ij} {=} \{ \bar{\mathbf{f}}_\text{DINO}(\mathbf{I}_s^{ij}){:}( \mu_s^{ij}, \sigma_s^{ij})\}_{i=1,\dots,N}^{j=1,\dots,M} \).
With this compact style dictionary, we can efficiently perform local style transfer using multiple reference images 
by semantically matching the clusters with the 3D scene.

\begin{figure}[t]
    \centering
    \includegraphics[width=\linewidth]{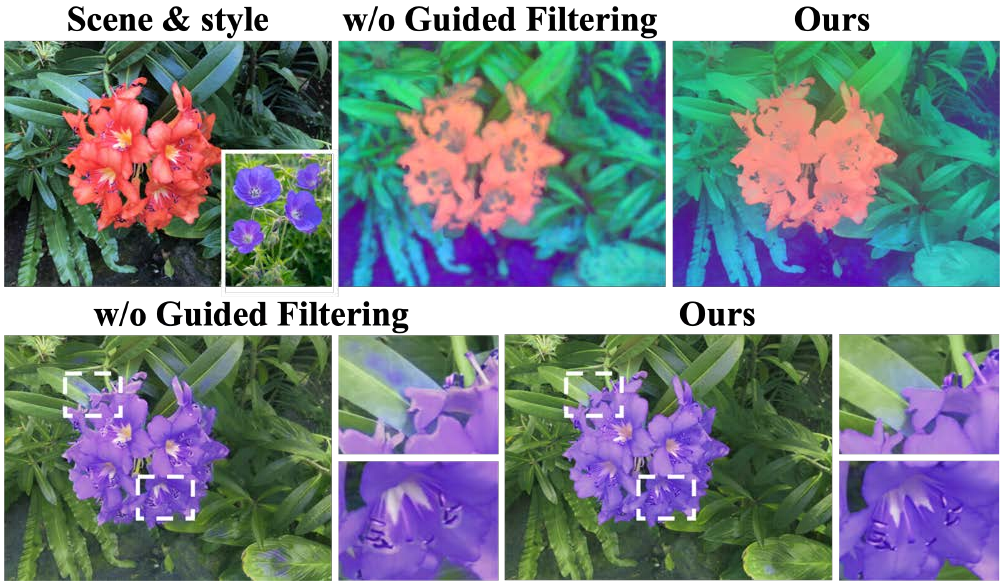} 
    \caption{\textbf{Effects of guided filtering on semantic features.} 
    \textbf{[Top]} Given 
    the trained 3D scene and the reference image (left), we visualize the
    learned stylizable radiance field without (mid) / with (right) guided filtering. 
    The learned semantic features are much sharper when guided filtering is applied.
    %
    \textbf{[Bottom]} The stylizable radiance field
    shows degraded 
    stylization results if learned without guided filtering, \eg, blurry boundaries (left), 
    higher stylization quality when learned with guided filtering (right).}
    \label{fig:guided_filtering_ablation}
\end{figure}

\paragraph{Scene semantic field}
%
To semantically match
the 3D scene and the reference image clusters,
we design and learn an auxiliary 
3D feature grid, called scene semantic field.
The scene semantic field contains semantic features of the 3D scene~\cite{kobayashi2022decomposing, tschernezki2022neural, kerr2023lerf}.
To distill semantic features to the scene semantic field, we optimize the tri-plane features and MLP (orange grid and MLP in \Fref{fig:pipeline_train}) by minimizing the error
between rendered features \(\hat{F}_\text{DINO}(\mathbf{r})\)
and features extracted from the input images by DINO as follows:
\begin{equation}
    \mathcal{L}_\text{semantic} = \sum_{\mathbf{r} \in \mathcal{R}} \lVert \hat{\mathbf{F}}_\text{DINO}(\mathbf{r}) - \mathbf{F}_\text{DINO}(\mathbf{I}, \mathbf{r})\rVert_1,
    \label{eq:loss_semantic}
\end{equation}
where \(\mathbf{F}_\text{DINO}(\mathbf{I}, \mathbf{r})\) denotes ground truth DINO features matched to ray \(\mathbf{r}\).
The density $\sigma$ from the scene content field is used for volume rendering, and \(\mathcal{L}_\text{semantic}\) does not affect the learning of the density.
We do not query view direction as input, in order to preserve
multi-view semantic consistency. 
Also, for constructing a fine-grained scene semantic field, the ground truth DINO feature maps are refined by guided filtering~\cite{he2012guided}.
The guided filtering enables consistent distillation of semantic features, resulting in clean and photorealistic stylized outputs (see \Fref{fig:guided_filtering_ablation}).

\paragraph{Semantic correspondence matching \& Local AdaIN}
When rendering the stylized 3D scene, we use two semantic feature matrices for computing semantic correspondence between the 3D scene and the elements of the style dictionary $\mathcal{D}$. 
The first one is 
\(\hat{\mathbf{F}}_\text{DINO} \in \mathbb{R} ^{K\times C_{D}}\), obtained from the scene semantic field, where \(K\) is the number of queried 3D points, and \(C_{D}\) is the channel size of the semantic feature, \ie, DINO feature.
The other one is \(\bar{\mathbf{F}}_\text{DINO}(\mathbf{I}_s) \in \mathbb{R} ^{T \times C_{D}} \) comprising keys \(\bar{\mathbf{F}}_\text{DINO}(\mathbf{I}_s^{ij})\) of 
\(\mathcal{D}\), which are the centroids of the \(T\) 
clusters.
Given
semantic feature matrices, \(\hat{\mathbf{F}}_\text{DINO}\) and \(\bar{\mathbf{F}}_\text{DINO}(\mathbf{I}_s)\), 
we compute a cross-correlation matrix \(\mathbf{R}\) as:
%
\begin{equation}
    \mathbf{R} = \hat{\mathbf{F}}_\text{DINO}\,\bar{\mathbf{F}}_\text{DINO}(\mathbf{I}_s)^{\top}.
    \label{eq:cross correlation}
\end{equation}

For mapping local styles of the reference images according to \(\textbf{R}\), we compose two matrices, \(\textbf{M}(\mathbf{F}_\text{VGG}(\mathbf{I}_s)) \in \mathbb{R} ^{T \times C_{V}}\) and \( \mathbf{\Sigma}(\mathbf{F}_\text{VGG}(\mathbf{I}_s)) \in \mathbb{R} ^{T \times C_{V}}\),
comprising of the local style codes (\(\mu_s^{ij}\), \(\sigma_s^{ij}\)) from the style dictionary,
%
where \(C_{V}\) denotes the channel size of VGG feature map.
%
We compute the matrix form of semantic-weighted style codes $(\textbf{M}_{w}, \mathbf{\Sigma}_w)$ as follows:
\begin{equation}
    \begin{gathered}
    \textbf{M}_w = \mathbf{R}^\texttt{S} \textbf{M}(\mathbf{F}_\text{VGG}(\mathbf{I}_s)) \\ \mathbf{\Sigma}_w = \mathbf{R}^\texttt{S} \mathbf{\Sigma}(\mathbf{F}_\text{VGG}(\mathbf{I}_s)),
    \label{eq:mw_sw}
    \end{gathered}
\end{equation}
where \(\mathbf{R}^\texttt{S} = \texttt{Softmax}(\mathbf{R})\) demonstrates style attention assigned to the queried 3D point features.
The semantic-weighted style code (\(\textbf{M}_w\), \(\mathbf{\Sigma}_w\))
are assigned to each 3D point according to the style attention.
We feed these semantic-weighted style codes to AdaIN layer as \( \mathbf{\Sigma}_w\sfrac{(\hat{\mathbf{F}}_\text{VGG} - \mu_c)}{\sigma_c} + \textbf{M}_w\), and perform volumetric rendering to obtain final stylized scene renderings.
Note that this semantic-aware local AdaIN preserves the multi-view stylized color consistency 
by directly measuring semantic correspondence between reference images and features on the scene semantic fields~\cite{kobayashi2022decomposing, kerr2023lerf}.
 

\begin{figure*}[thbp]
    \centering
        \includegraphics[width=\linewidth]{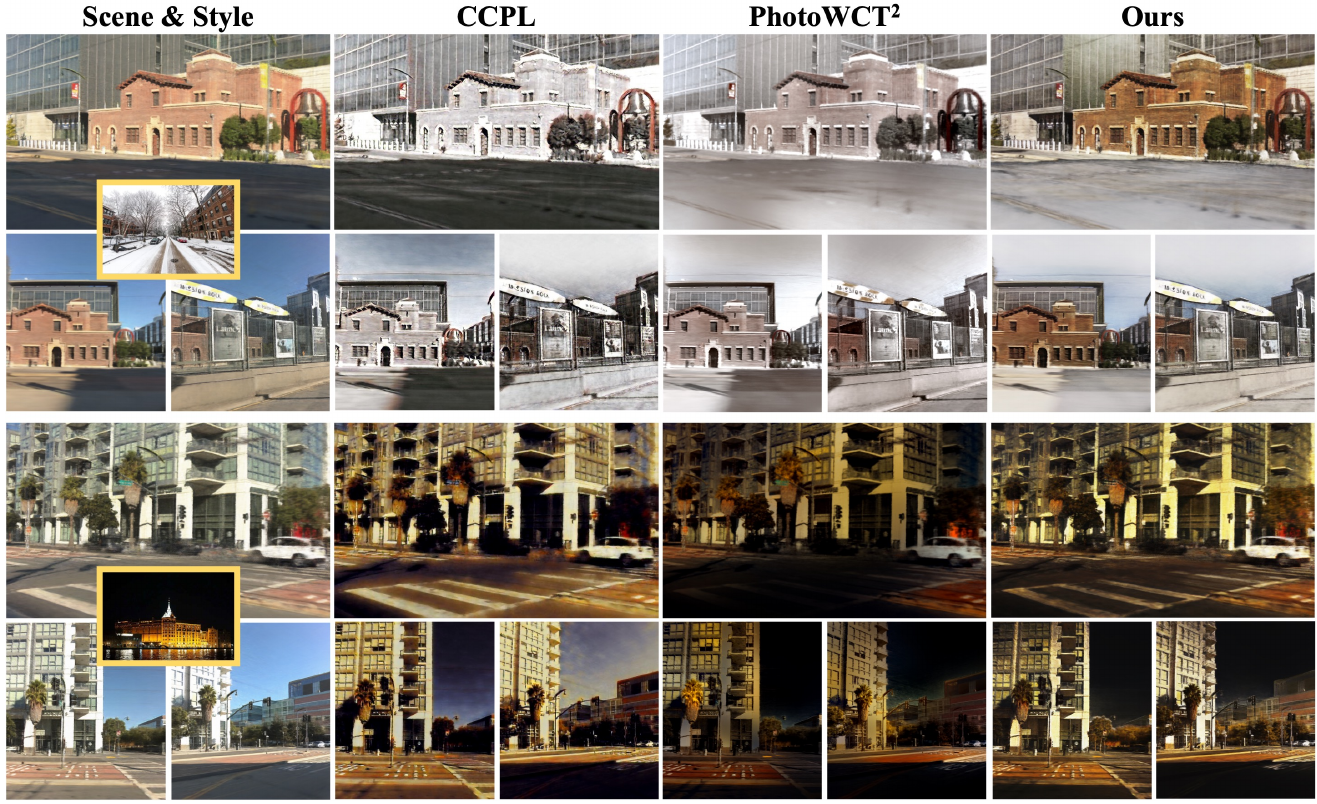}
    \caption{
   \textbf{Multi-view appearance consistency on the San Francisco Mission Bay dataset~\cite{tancik2022blocknerf}.}
   \ours~preserves multi-view appearance consistency even in extreme viewpoint change, while 2D PST methods (\citet{wu2022ccpl}; ~\citet{chiu2022photowct2}) produce inconsistent colors of the same building as the viewpoint changes. }
    \vspace{3mm}
\label{fig:large_scale_comparison}
        \centering
        \includegraphics[width=\linewidth]{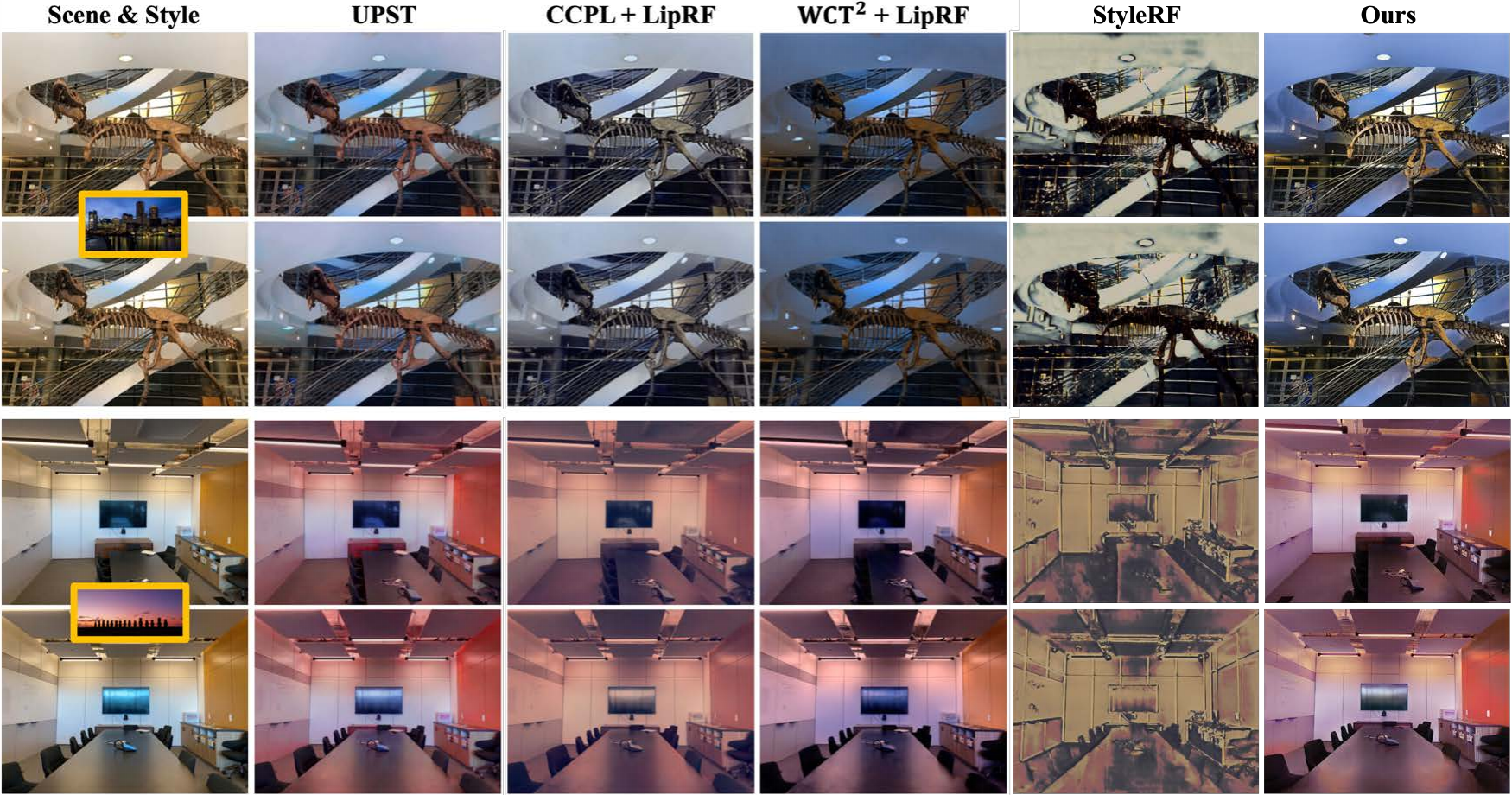}
        \caption{\textbf{PST quality comparison on the LLFF dataset~\cite{mildenhall2019local}.}
        Compared to the competing 3D PST methods, ~\ours~ stylizes the radiance field in a photorealistic manner by transferring the diverse color of the reference image while preserving the original images’ naturalness and vividness.
.
        }
\label{fig:LLFF_comparison}
\end{figure*}

We found that
%
computing \(\textbf{M}_w\) and \(\mathbf{\Sigma}_w\) on the feature map resolution without clustering~\cite{gunawan2023modernizing}
is inefficient for the volume rendering framework where 
iterative rendering of rays is inevitable.
Our clustering enables efficient style transfer, especially for cases using multiple reference images.
We highlight that photorealistic scene stylization results are obtained
when ${\sim}10$ clusters are used. 
It is worth noting that 
the clustering process takes no more than 1 \emph{sec.} for each reference image. 
Using the small number of clusters allows us to avoid iterative concatenation of high-dimensional reference image features 
and effectively reduces the computational cost of matrix multiplication.



\section{Experiments}
\label{sec:experiment}
In this section, we demonstrate the 
qualitative results of FPRF on large- and small-scale scenes, and ablation studies.
%




\begin{figure}[t]
    \centering
    \includegraphics[width=1\linewidth]{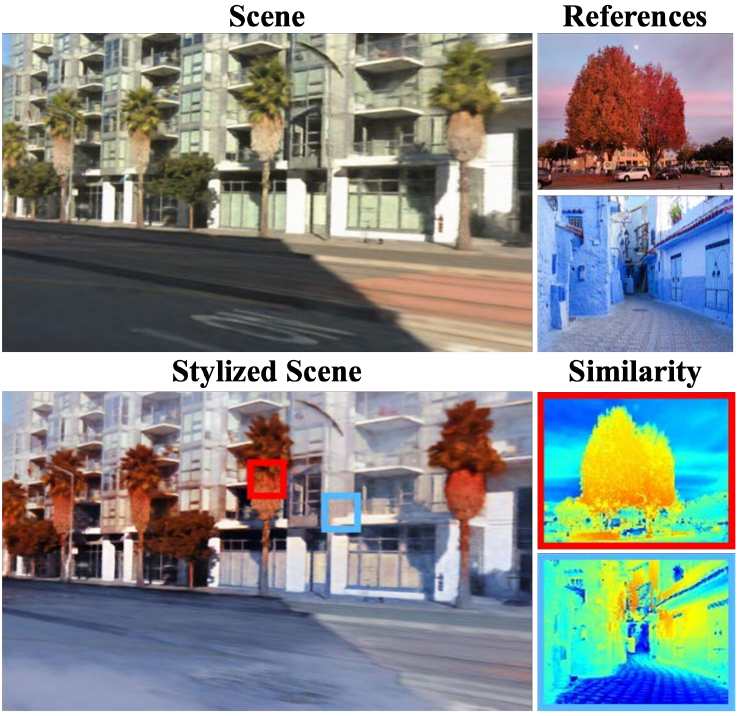} 
    \caption{\textbf{Multi-reference style transfer.}
    \ours~stylizes the 3D radiance field with multiple reference images. Each heatmap shows the similarity between the 
    semantic features of a highlighted patch and the reference image. Our model comprehends the 
    semantic relationship of a large-scale 3D scene and 
    matches the scene with the reference images.
    }
    \label{fig:multi_reference}
\end{figure}

\paragraph{PST on large-scale scenes}
For large-scale scenes, we consider the San Francisco Mission Bay dataset~\cite{tancik2022blocknerf}, a city scene dataset consisting of about 12,000 images recorded by 12 cameras.
Since we propose the first method aiming for large-scale 3D scene PST, no 3D PST method supports large-scale datasets.
Therefore, we compare FPRF against two competing 2D PST methods, CCPL~\cite{wu2022ccpl} and \(\text{PhotoWCT}^2\)~\cite{chiu2022photowct2}.
Figure~\ref{fig:large_scale_comparison} illustrates a qualitative comparison on large-scale scenes.
%
%
We observe
that 2D PST methods fail to preserve multi-view color 
consistency under wide-range view changes, \eg,
they obtain 
different colors of the same building and sky as the viewpoints change.  
Also, they 
stylize images without understanding of semantic relation between the content image and the reference image.
Instead, our FPRF elaborately transfers styles by reflecting the semantic correspondence between scene and reference images.
Semantic matching can also preserve multi-view consistency by directly measuring semantic correspondence between reference images and 3D points in the 3D scene semantic field directly. 
%

\begin{figure}[t]
    \centering
    \includegraphics[width=\linewidth]{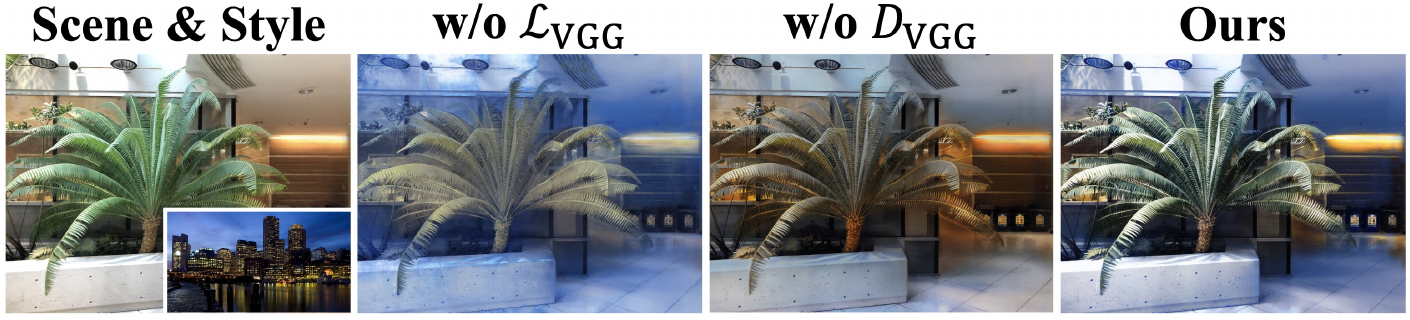}
    \caption{\textbf{Ablation studies for the VGG feature distillation loss and the pre-trained color decoder.} ``w/o $\mathcal{L}_\text{VGG}$'' trains model without the VGG feature distillation loss  $\mathcal{L}_\text{VGG}$. ``w/o $D_\text{VGG}$'' replaces the pre-trained color decoder $D_\text{VGG}$ to an MLP which is trained from scratch during scene optimization. 
    }
    \label{fig:vgg_ablation}
\end{figure}

\paragraph{PST on small-scale scenes}
For small-scale scenes, we consider LLFF~\cite{mildenhall2019local}, which includes eight real forward-facing scenes.
We compare our model with two 
competing
3D PST methods; LipRF~\cite{zhang2023transforming}, UPST-NeRF~\cite{chen2022upst}, and 
the most recent
feed-forward 3D 
\emph{artistic}
style transfer method, StyleRF~\cite{liu2023stylerf}.
As shown in Fig.~\ref{fig:LLFF_comparison}, our method transfers the diverse colors of the style reference image well while maintaining the 
texture fidelity of the original scene.
While UPST-NeRF effectively retains the 
structure of the scene, the stylized scene's color style differs from the reference.
LipRF stabilizes the artifacts caused by 2D PST methods with the Lipschitz network, but it tends to oversmooth the scene and loses the 
color diversity of the reference image.
%
Noticeably, StyleRF fails to obtain photorealistic results and loses the detailed structure of the original scene.
Note that UPST-NeRF and StyleRF require per-scene optimization after scene reconstruction for style transfer, and LipRF needs per-style optimization. 
On the contrary, our model achieves feed-forward PST after an efficient single-stage learning of the stylizable radiance field.

\paragraph{Multi-reference style transfer}
Figure~\ref{fig:multi_reference} shows the style transfer result using multiple reference images. 
Our model effectively selects suitable styles from multiple references with scene semantic field. 
Semantic similarity is computed by multiplying features from the scene semantic field with DINO~\cite{caron2021emerging} feature maps extracted from reference images. 
The similarity map clearly shows that our model comprehends the accurate semantic relationship between scenes and reference images.
 
\paragraph{Ablation studies}
Figure \ref{fig:vgg_ablation} shows the effects of the VGG feature distillation loss and the pre-trianed color decoder.
The VGG feature distillation loss helps preserve the original scene content and improve the quality of style transfer, by guiding the content feature to follow VGG feature distribution. The generalizability of the pre-trained color decoder allows the model to get arbitrary style reference images as input.






\section{Conclusion}
In this paper, we present \ours, a novel stylizable 3D radiance field 
aiming large-scale 3D scene photorealistic style transfer (PST). 
\ours~allows feed-forward 
PST after only a single-stage training by leveraging AdaIN.
%
%
FPRF also supports multi-reference style transfer, which allows stylizing large-scale 3D scenes which consist of diverse components.
%

The current limitation is that the semantic matching performance of our model is bounded by the capability of the semantic image encoder, DINO.
Nonetheless, 
since our model
can utilize any semantic encoder for constructing the scene semantic field, the performance of our model stands to benefit from the emergence of more advanced models.
%
%
%

\section*{Acknowledgments} This work was supported by the LG Display (2022008004), Institute of Information \& communications Technology Planning \& Evaluation (IITP) grant funded by the Korea government(MSIT) (No.2022-0-00124, Development of Artificial Intelligence Technology for Self-Improving Competency-Aware Learning Capabilities; No. RS-2022-00164860, Development of human digital twin technology based on dynamic behavior modeling and human-object-space interaction; No.2021-0-02068, Artificial Intelligence Innovation Hub; No. 2019-0-01906, Artificial Intelligence Graduate School Program(POSTECH)), and National Research Foundation of Korea (NRF) grant funded by the Korea government (MSIT) (No. NRF-2021R1C1C1006799).

\bibliography{ms}

\end{document}


\setcounter{section}{0}
\setcounter{figure}{0}
\setcounter{table}{0}
\setcounter{equation}{0}
\renewcommand\thesection{\Alph{section}}
\renewcommand\thefigure{S\arabic{figure}}
\renewcommand{\thetable}{S\arabic{table}}
\renewcommand\theequation{\alph{equation}}

\maketitle

This supplementary material aims to provide additional contents and details not included in the main paper due to space limitations. 
%
%
We provide 
additional experiments and results in \Sref{sec:supp_exp}. 
In \Sref{sec:supp_discussion}, we provide further discussion about the multi-view consistency of FPRF.
The separate supplementary video includes
the explanation of 
our key idea
and the video result demos 
for our 3D PST results.

\section{Experiment}
\label{sec:supp_exp}
In this section, we demonstrate the quantitative results, including scene reconstruction quality of FPRF, training/running time, and user study.

\subsection{Scene-agnostic pre-trained color decoder}
\label{scene_reconstruction}
FPRF is trained with the pre-trained MLP color decoder \(D_\text{VGG}\) compatible with AdaIN. 
This decoder is agnostic to both scenes and styles.
This decoder is trained on various independent images, which are not necessary to be from multiple-view images of a scene but just a separate large image dataset.
Thereby, it can be generalized across different scenes, regardless of the color distribution of the scene. 
To assess the quality of this decoder and understand the upper bound of the performance of our FPRF, we report
original scene reconstruction quality of FPRF in \Tref{table: scene reconstruction}.
%
The original scene is the rendered 3D radiance field without stylization, and the baseline denotes a model that shares the same architecture as FPRF but is trained with a learnable decoder from scratch, \ie, a scene-specialized decoder.
Although we train FPRF with a fixed pre-trained decoder agnostic to scenes and styles, it shows the comparable rendering quality with the scene-specialized baseline model.


\subsection{Training time}

\paragraph{Comparison with feed-forward methods}
\Tref{tab:supp_training_time}-\colorref{(a)} shows the comparison of running times.
In the table, FPRF shows that the least training time for scene reconstruction with neural radiance fields among the competing feed-forward methods.
This training time is required just once for a scene; thus, one can stylize with arbitrary styles without any additional optimization or training at all.
This advantage is achieved by 
our efficient single-stage training that leverages 
the pre-trained
scene-agnostic decoder.
In contrast, 
UPST-NeRF~\cite{chen2022upst} and StyleRF~\cite{liu2023stylerf} require time-consuming training processes even after the scene reconstruction to enable feed-forward stylization.
That is, these methods take longer scene reconstruction and require additional processing time not reported in the table, while our method does not.

\paragraph{Comparison with an optimized-based method}
We compare with an optimization-based method in
\Tref{tab:supp_training_time}-\colorref{(b)}.
In the table, we hold an advantage in our ability to perform multiple stylizations without additional optimization.
Different from the aforementioned feed-forward methods, LipRF~\cite{zhang2023transforming} requires per-scene optimization followed by per-style optimization to stylize 3D scenes. 
%
Note that, although the running time of 7 min.~per style appears to be reasonable, 
it is the time of tiny scale scenes like LLFF~\cite{mildenhall2019local}.
Thus, such additional per-style optimization is
a significant drawback and makes the approach inapplicable in 
stylizing large-scale scene radiance fields, typically decomposed into multiple radiance field blocks~\cite{tancik2022blocknerf}.
%


\begin{table}[t]
		\small
		\setlength\tabcolsep{1.5pt}
            \renewcommand{\arraystretch}{1.2}
		\begin{tabular}{c}
            \includegraphics[width=\linewidth]{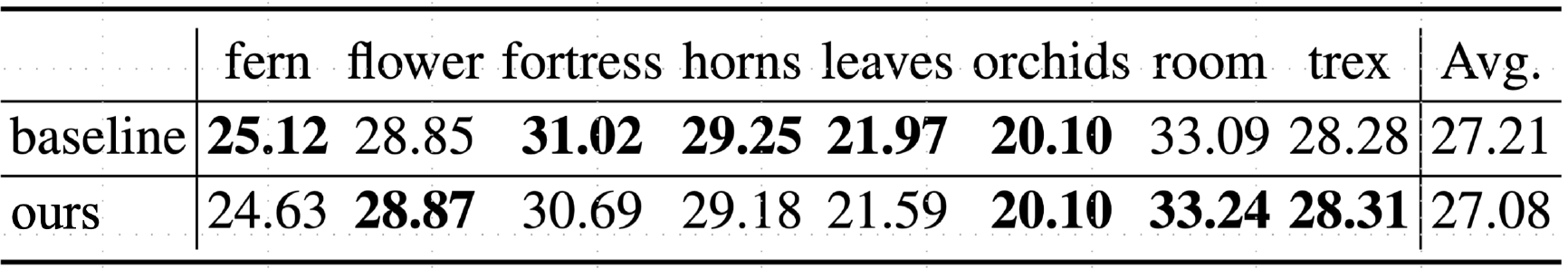}
		\end{tabular}
		\vspace{-0.3cm}
		\caption{\textbf{Comparison with the baseline model.}
                We compare the rendering quality of FPRF with a baseline model on the LLFF dataset.
                The baseline denotes a model 
                that has the same architecture as FPRF, but is trained with a learnable decoder from scratch.
                FPRF shows comparable PSNRs with the baseline. 
  }
		\label{table: scene reconstruction}
		\vspace{-1em}
	\end{table}

\begin{table}[t]
\centering
    \resizebox{\linewidth}{!}{
    \begin{tabular}{l ccc}
        \toprule
        & {UPST-NeRF} & {StyleRF} &  {FPRF (Ours)}\\

        \midrule

        scene recon.~time & $\approx$ 10 hrs. &  $\approx$ 6 hrs. & $\approx$ 1 hr. \\

        \bottomrule
    \end{tabular}
    }
    (a) Comparison to feed-forward methods.\vspace{2mm}
    \resizebox{\linewidth}{!}{
    \begin{tabular}{l cc}
        \toprule
        &  {LipRF} &  {FPRF (Ours)}\\

        \midrule

        stylization time
        &  $\approx$
        7 min./per-scene & none
        \\
        \bottomrule
    \end{tabular}
    }
    \vspace{1mm}(b) Comparison to an optimization-based method.
    \caption{\textbf{Running time comparisons.
    } (a) Comparison with feed-forward 3D style transfer methods.
    FPRF requires less training time than 
    the other methods,
    and achieves this efficiency
    just through an efficient single-stage learning process.
    (b) Comparison with an optimization-based PST method.
    FPRF does not require a separate optimization for stylization but just
    feed-forwarding, while LipRF~\cite{zhang2023transforming} requires per-style and per-scene optimization.
    }\vspace{-3mm}
    \label{tab:supp_training_time}
\end{table}





\begin{figure*}[t]
    \centering
    \includegraphics[width=\linewidth]{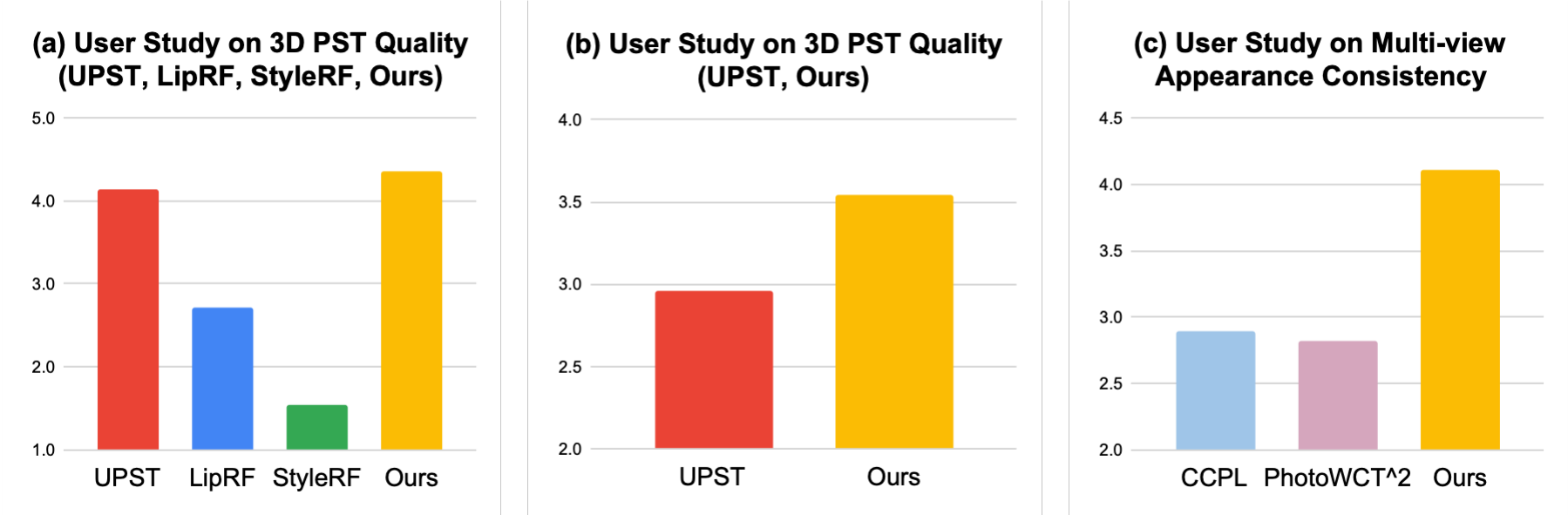} 
    \caption{ \textbf{User study results}
    \textbf{(a)} Comparison with 3D style transfer methods on LLFF~\cite{mildenhall2019local} dataset. \textbf{(b)} Comparison with UPST-NeRF on LLFF dataset using more samples. \textbf{(c)} Comparison with 2D PST methods on San Francisco Mission Bay dataset~\cite{tancik2022blocknerf}.
     FPRF outperforms competing 3D methods in stylization quality, and surpasses 2D methods multi-view consistency.
     }\vspace{-3mm}
    \label{fig:supp_user_study}
\end{figure*}

\subsection{User study}
\label{user_study}

We conduct a user study to evaluate the 
perceptual quality of stylization.
We evaluate FPRF over the LLFF~\cite{mildenhall2019local} dataset for the small-scale scene and the San Francisco Mission Bay dataset~\cite{tancik2022blocknerf} for the large-scale scene. We asked 30 volunteers to score (1-5) the stylization quality of 3D PST in small scenes and multi-view consistency for large-scale scenes.
%

\begin{figure*}[thbp]
    \centering
        \includegraphics[width=\linewidth]{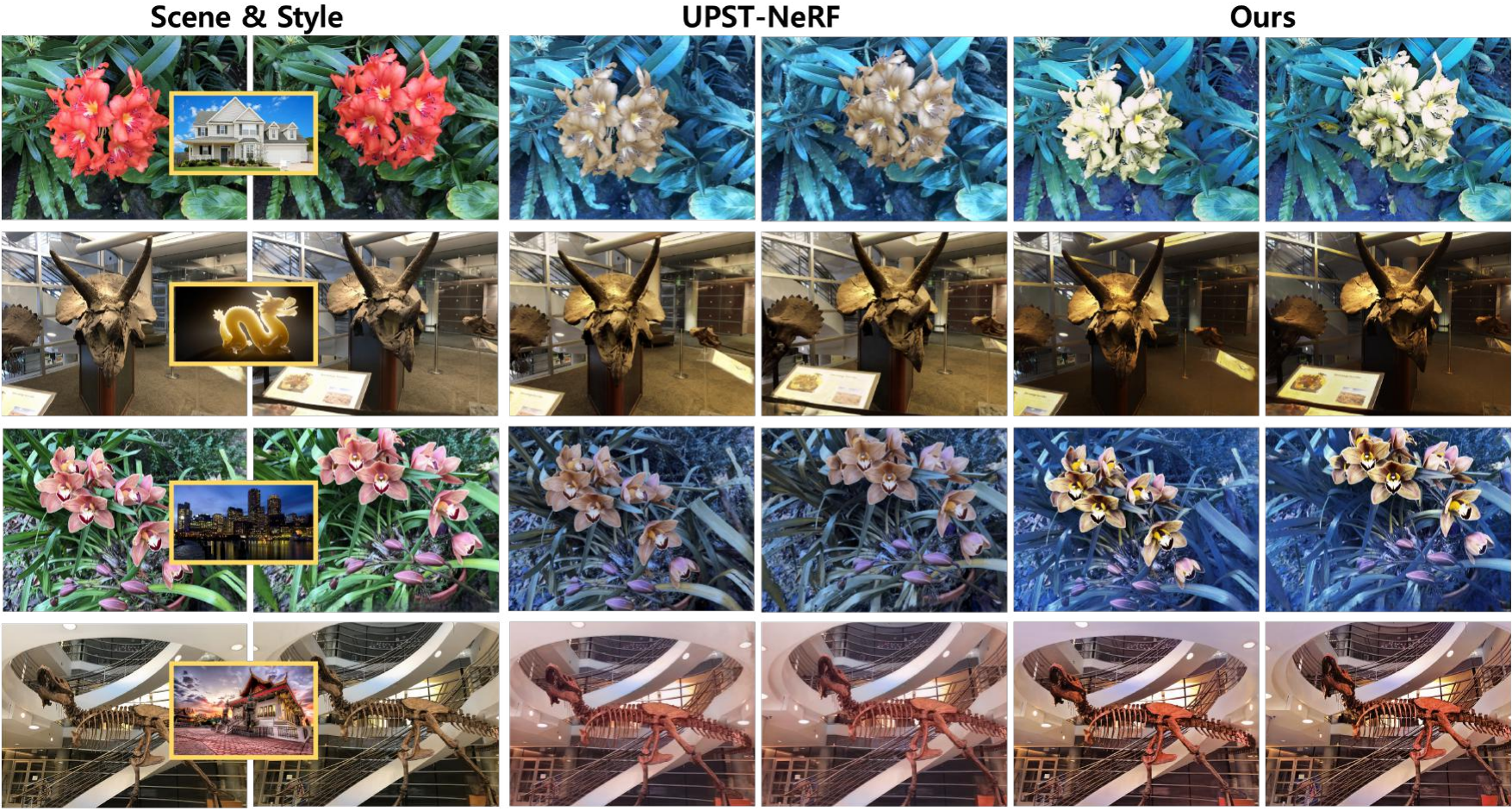}
                \caption{\textbf{Additional qualitative results on LLFF dataset~\cite{mildenhall2019local}.}
        Compared to the UPST-NeRF~\cite{chen2022upst}, FPRF accurately reflects the diverse color of the reference image.}\vspace{3mm}
        
\label{fig:supp_LLFF}
    \centering
        \includegraphics[width=\linewidth]{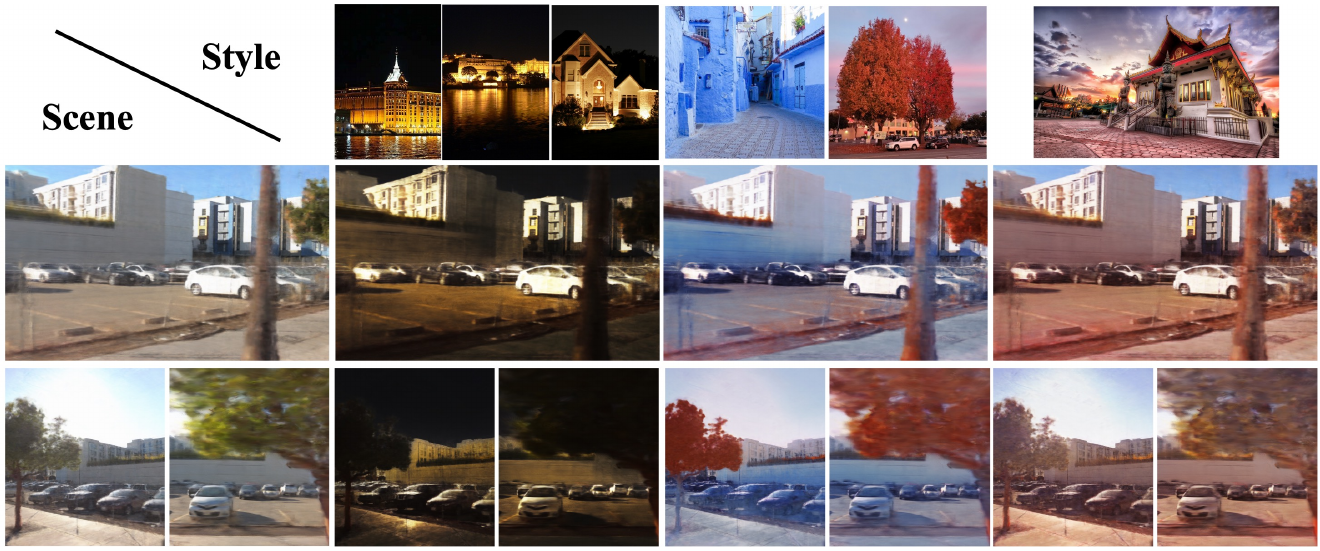}
        \caption{   \textbf{Additional qualitative results on San Francisco Mission Bay dataset~\cite{tancik2022blocknerf}.}
           FPRF performs high-quality style transfer using multi-reference style images through semantic matching.
        }
    \vspace{3mm}
\label{fig:supp_large}

\end{figure*}

\paragraph{LLFF dataset}
We compare FPRF with other 3D PST methods, UPST-NeRF~\cite{chen2022upst}, LipRF~\cite{zhang2023transforming}, and an artistic style transfer method StyleRF~\cite{liu2023stylerf} on the LLFF dataset. 
%
%
We further conduct the second study comparing with UPST-NeRF using eight pairs of scenes and style images.
As shown in \Fref{fig:supp_user_study}-\colorref{(a),(b)},
FPRF gets higher scores than the competing 3D PST methods in terms of the quality of stylization.
Please refer to \Fref{fig:supp_LLFF} for the samples used.
%

\paragraph{San Francisco Mission Bay dataset}
For large-scale scenes, we compare FPRF with competing 2D PST methods, CCPL~\cite{wu2022ccpl} and $\text{PhotoWCT}^2$~\cite{chiu2022photowct2}, since there exists 
no 3D PST method that supports large-scale 3D radiance fields.
Figure \ref{fig:supp_user_study}-\colorref{(c)} shows that FPRF outperforms 2D PST methods in terms of multi-view consistency.
Please refer to \Fref{fig:supp_large} for the large-scale scene samples we used.

\section{Is FPRF Multi-view Consistent?}
\label{sec:supp_discussion}

\label{multi_view}
The main challenge of 3D scene style transfer is to preserve 
multi-view consistency after stylization. 
Multi-view consistency denotes a feature
that a shared geometry wherein a scene observed from various angles can consistently explain multiple views of the scene.
FPRF 
effectively preserves
multi-view consistency by computing semantic correspondence with features on the 3D space directly, not on the 
rendered features like \citet{kobayashi2022decomposing, kerr2023lerf}.
Our scene semantic field gets only the 3D positions as input, without the viewing direction, and outputs the 
consistent
semantic features of the 3D position.
This consistent 3D semantic feature
ensures that each 3D point is always stylized with the assigned same local style, regardless of 
view directions. 
Note that FPRF can also reflect 
non-Lambertian effects
with the scene content field trained with the view directions as inputs.
%
%

Unfortunately, this favorable and advantageous property cannot be effectively assessed by the existing consistency metric.
A common metric for evaluating the multi-view consistency 
of 3D stylization is to compute the multi-view error via image warping~\cite{lai2018learning}. 
Specifically, for the two synthesized 
images \(\textbf{I}_\textbf{d} \) and \(,\textbf{I}_{\textbf{d}'}\) from two different views \(\textbf{d}\) and \(\textbf{d}'\), the multi-view error is computed as follows:
\begin{equation}
    E_\text{warp}(\textbf{I}_\textbf{d}, \textbf{I}_{\textbf{d}'}) = \text{MSE}(\textbf{I}_\textbf{d}, W(\textbf{I}_{\textbf{d}'}); \textbf{B}_{\textbf{d}'\textbf{d}}),
\end{equation}
where \(W\) warps \(\textbf{I}_{\textbf{d}'}\) to \(\textbf{I}_\textbf{d}\) and $\textbf{B}_{\textbf{d}'\textbf{d}}$ is the binary mask of valid pixels warped from the view \(\textbf{d}'\) to \(\textbf{d}\). The warping function and binary mask can be obtained by predicted depth maps of the rendered images or be measured from an off-the-shelf pre-trained optical flow method~\cite{teed2020raft}. The non-valid pixels are masked out and are
not considered. 
This proposed warp error tries to quantify the difference in RGB values between images rendered with different views.
However, this
metric is limited in that
it can be significantly affected by material property 
according to
the change of view directions, \ie, view-dependent reflectance.
Moreover, this metric can yield a low error for over-smoothed outcomes, regardless of the fidelity of the resulting image as pointed out by \citet{liu2023stylerf}.
Hence, we do not employ this metric due to its inability to quantify true multi-view consistency for non-Lambertian effects, which our method can deal with.

\bibliography{supplement}